%% file: main-with-authors.tex
\documentclass[a4paper,fleqn]{cas-sc}
\usepackage[nomath]{lmodern}
\usepackage{apacite}
\usepackage[authoryear]{natbib}
\usepackage{url}
\usepackage{multirow}
\usepackage{booktabs}
\usepackage{tcolorbox}
\usepackage{subcaption}
\usepackage{caption}
\usepackage{amssymb}
\usepackage{amsmath}
\usepackage{graphicx}
\usepackage{xcolor}
\usepackage{colortbl}
\usepackage{longtable}
\usepackage{rotating}
\usepackage{xspace}
\usepackage{hyperref}
\definecolor{crossrefcolor}{RGB}{35,80,115}
\definecolor{DarkSlateGrey}{RGB}{47,79,79}
\hypersetup{
  colorlinks=true,
  linkcolor=crossrefcolor,
  citecolor=black,
  urlcolor=DarkSlateGrey,
  filecolor=crossrefcolor
}
\usepackage[nameinlink,noabbrev]{cleveref}
\crefname{section}{Section}{Sections}
\Crefname{section}{Section}{Sections}
\crefname{subsection}{\S}{\S\S}
\Crefname{subsection}{\S}{\S\S}
\crefname{subsubsection}{\S}{\S\S}
\Crefname{subsubsection}{\S}{\S\S}
\crefname{appendix}{Appendix}{Appendices}
\Crefname{appendix}{Appendix}{Appendices}
\crefname{figure}{Figure}{Figures}
\Crefname{figure}{Figure}{Figures}
\crefname{table}{Table}{Tables}
\Crefname{table}{Table}{Tables}
\newcommand{\appref}[1]{\hyperref[#1]{Appendix~\ref*{#1}}}
\definecolor{gain}{rgb}{0.85, 1, 0.85}
\definecolor{loss}{rgb}{1, 0.85, 0.85}
\definecolor{gaintext}{RGB}{0,112,60}
\definecolor{losstext}{RGB}{192,0,0}
\newtcolorbox{grayquote}{colback=gray!10,colframe=white,boxrule=0pt,left=5pt,right=5pt,top=5pt,bottom=5pt}

\begin{document}
\let\WriteBookmarks\relax
\def\floatpagepagefraction{1}
\def\textpagefraction{.001}

\shorttitle{CAR}
\shortauthors{Song et al.}
\input{author-title-block}
\input{abstract}
\input{keywords}
\ExplSyntaxOn
\bool_gset_true:N \g_stm_nologo_bool
\ExplSyntaxOff
\maketitle
\input{body}
\clearpage
\renewcommand{\bibliographytypesize}{\fontsize{8pt}{9.5pt}\selectfont}
\renewcommand{\bibfont}{\fontsize{8pt}{9.5pt}\selectfont}
\bibliographystyle{apacite}
\bibliography{reference}
\end{document}

%% file: author-title-block.tex
\shorttitle{CAR}
\shortauthors{Song et~al.}
\title [mode = title]{CAR: Query-Guided Confidence-Aware Reranking for Retrieval-Augmented Generation}

\author[dut]{Zhipeng~Song}[orcid=0009-0009-6249-1988]
\ead{songzhipeng@mail.dlut.edu.cn}
\author[dlou]{Yizhi~Zhou}[orcid=0000-0002-6761-5953]
\ead{zhouyizhi@dlou.edu.cn}
\author[ldu]{Xiangyu~Kong}[orcid=0000-0003-1940-8674]
\ead{xiangyukong@liaodongu.edu.cn}
\author[dut,qhu]{Jiulong~Jiao}[orcid=0009-0001-9852-7999]
\ead{jiaojiulong@mail.dlut.edu.cn}
\author[dut]{Xuezhou~Ye}[orcid=0009-0002-3421-5889]
\ead{yexzh6@mail2.sysu.edu.cn}
\author[dut]{Chunqi~Gao}[orcid=0009-0008-8576-3141]
\ead{gaochunqi@mail.dlut.edu.cn}
\author[dmu]{Xueqing~Shi}[orcid=0009-0008-3754-6396]
\ead{shixq@dmu.edu.cn}
\author[hit]{Yu~Wang}[orcid=0000-0003-3008-8712]
\ead{ywang@nagoya-u.jp}
\author[tencentdl]{Yuhang~Zhou}[orcid=0009-0007-3489-0549]
\ead{ginozhou@tencent.com}
\author[dut]{Heng~Qi}[orcid=0000-0002-8770-3934]
\cormark[1]
\ead{hengqi@dlut.edu.cn}
\cortext[cor1]{Corresponding author.}

\affiliation[dut]{organization={School of Computer Science and Technology, Dalian~University~of~Technology}, addressline={No.2 Linggong Road, Ganjingzi District}, city={Dalian}, postcode={116024}, country={China}}
\affiliation[dlou]{organization={School of Information Engineering, Dalian Ocean University}, addressline={No. 2-52, Heishijiao Street, Shahekou District}, city={Dalian}, postcode={116023}, country={China}}
\affiliation[ldu]{organization={School of Information Engineering, Liaodong~University}, addressline={No.116 Linjiang Back Street, Zhenan District}, city={Dandong}, postcode={118001}, country={China}}
\affiliation[qhu]{organization={Information Technology Center, Qinghai~University}, addressline={251 Ningda Road, Chengbei District}, city={Xining}, postcode={810016}, country={China}}
\affiliation[dmu]{organization={College of Health-Preservation and Wellness, Dalian Medical University}, addressline={No. 9 West Section of Lvshun South Road, Lvshunkou District}, city={Dalian}, postcode={116044}, country={China}}
\affiliation[tencentdl]{organization={Tencent (Dalian Northern Interactive Entertainment Technology Co., Ltd.)}, addressline={21/F, Tencent Building, No. 26 Jingxian St, Ganjingzi District}, city={Dalian}, postcode={116085}, country={China}}
\affiliation[hit]{organization={Graduate School of Social Data Science, Hitotsubashi University}, addressline={2-1 Naka, Kunitachi}, city={Tokyo}, postcode={1868601}, country={Japan}}

%% file: abstract.tex
\begin{abstract}
Retrieval-augmented generation (RAG) relies on evidence ranking to determine what information is exposed to the generator, yet existing retrieval and reranking methods primarily estimate query--document relevance. Relevance, however, is not equivalent to generator-side usefulness: a relevant passage may introduce ambiguity or distraction, whereas a lower-ranked passage may stabilize the generator's answer. We present CAR (Confidence-Aware Reranking), a training-free rank-correction framework that uses query-only answer stability as a control and measures each candidate by the change it induces in sampled-answer semantic stability. This controlled contrast estimates a document's marginal contribution to generator behavior without treating semantic stability as relevance or calibrated correctness. CAR converts these confidence changes into coarse precedence constraints and returns the feasible ranking with minimum Kendall distance from the baseline, preserving existing pairwise preferences unless generator-side evidence supports reversing them. Experiments on NQ, HotpotQA and FEVER across sparse and dense retrievers, seven ranking methods and three generator families show robust improvements. In the BM25-centered main analysis, CAR achieves a \textbf{+5.53\% mean relative NDCG@5 gain}; on the fixed NQ-answerable downstream evaluation, it improves token-level F1 by \textbf{+0.43 points}, with ranking and generation gains strongly aligned across rankers ($\rho = 0.93$). These results position CAR as a deployment-friendly, generator-aware correction layer that complements relevance while preserving informative prior rankings. CAR requires neither task-specific training nor access to model internals such as logits or hidden states, making it applicable to black-box LLMs through generated outputs alone.
\end{abstract}

%% file: keywords.tex

%% file: body.tex
\section{Introduction}
\label{sec:introduction}

Retrieval-augmented generation (RAG) reduces unsupported generation by conditioning large language models (LLMs) on external evidence \citep{rag,gao2023rag}. This design has become central to open-domain question answering, fact verification and domain-specific knowledge access \citep{ipm-1} because it shifts part of the burden from parametric memory to retrieved documents. Yet a RAG generator usually receives only a small set of top-ranked passages. The retrieval and reranking stages therefore determine both perceived relevance and the evidence available for the final answer.

Most retrieval and reranking methods are optimized around query-document relevance. Sparse retrieval emphasizes lexical matching, dense retrieval learns representation-level matching, supervised neural rerankers learn fine-grained relevance from labeled data, and LLM-based rerankers use prompts or likelihood signals to reorder candidate documents \citep{ipm-2}. These relevance signals are valuable, but they are not identical to generation usefulness. A passage may be lexically or semantically related to the query while still introducing ambiguity, distraction or conflicting evidence for the generator. Conversely, a lower-ranked passage may provide the missing evidence that makes the generator's answer more stable.

This gap matters because retrieval ultimately serveeliable answer generation. A useful passage should make the generator's sampled answer meanings more stable under the current query. Repeated sampling from an LLM provides an observable signal for this purpose: when the model has enough evidence, sampled answers tend to be semantically consistent; when evidence is weak or conflicting, sampled answers tend to spread across different semantic clusters \citep{wang2023selfconsistency,farquhar2024detecting}. This suggests that the change in sampled-answer consistency after adding a document can reveal whether the document is useful to the generator.

We propose \textbf{CAR} (\textbf{C}onfidence-\textbf{A}ware \textbf{R}eranking), a query-guided and training-free rank-correction framework that complements relevance-based ranking with generator-side answer stability. CAR contrasts query-only and document-conditioned confidence to estimate each document's marginal contribution to semantic answer stability. It converts this evidence into coarse precedence constraints and returns the feasible ranking closest to the baseline in Kendall distance, thereby limiting unnecessary reversals of the existing order. Because CAR requires only sampled generator outputs and an existing candidate ranking, it supports black-box integration without access to model internals or retraining.

\begin{center}
\centering
\input{wrappers/fig1_framework_overview_en}
\captionof{figure}{\textbf{Overview of the CAR Framework.} Starting from an existing candidate ranking, CAR estimates query-only semantic answer stability and opens correction only when it falls below QT. It then compares document-conditioned stability with the query-only control, assigns documents to promote, preserve or demote bins using CM, and retains the baseline order within each bin. The resulting ranking incorporates generator-side evidence while limiting unnecessary changes to the original order.}
\label{fig:car_framework}
\end{center}

The main contributions of this work are threefold:

\begin{itemize}
    \item \textbf{Generator-oriented, query-conditioned usefulness signal.}
    We introduce a generator-oriented, query-conditioned usefulness signal that uses query-only confidence as a control and characterizes the relative change in document-conditioned confidence as a document's marginal contribution to the generator's semantic answer stability.

    \item \textbf{Deployment-friendly conservative rank correction.}
    We present a training-free rank-correction framework that integrates directly into existing retrieval and reranking pipelines. CAR uses only sampled generator outputs and an existing candidate ranking, requiring neither access to model internals nor retraining of the generator or baseline ranker. Its order-preserving design retains the baseline order wherever generator feedback does not support a change, thereby limiting unnecessary rank reversals and reducing the risk of overcorrection.

    \item \textbf{Robustness across retrieval and generation settings.}
    We evaluate CAR on three datasets spanning open-domain question answering, multi-hop question answering, and fact verification, using sparse and dense retrievers, seven baseline ranking methods, and multiple generator families. The results show robust overall gains across these varied settings, while sensitivity analyses demonstrate that CAR remains effective over a broad range of parameter values.
\end{itemize}

The remainder of this paper is organized as follows. \cref{sec:related_work} reviews related work on RAG evidence selection, reranking and confidence estimation, and positions CAR within this literature. \cref{sec:methodology} presents the formulation and rank-correction algorithm of CAR. \cref{sec:experiments} evaluates its ranking effectiveness, component contributions, operating robustness and downstream answer-quality consistency. \cref{sec:conclusion} summarizes the findings and discusses their implications, limitations and directions for future work.

\section{Related Work}
\label{sec:related_work}

\subsection{RAG and Evidence Selection}
Retrieval-augmented generation couples parametric generation with access to external evidence \citep{rag,gao2023rag}. Early and large-scale formulations establish several ways to realize this coupling. REALM learns a latent retriever jointly with language-model pretraining, allowing evidence to be consulted during pretraining, fine-tuning and inference \citep{pmlr-v119-guu20a}. RETRO conditions autoregressive prediction on chunks retrieved from a very large text database \citep{pmlr-v162-borgeaud22a}, whereas Atlas integrates retrieval with few-shot learning over an updatable document index \citep{JMLR:v24:23-0037}. Although their architectures differ, these systems share a foundational premise: the information exposed to the generator is partly determined by retrieval rather than fixed parametric memory alone.

First-stage retrieval determines which documents are available for subsequent reasoning and ranking. BM25 represents lexical matching, Dense Passage Retrieval learns supervised dual-encoder retrieval for open-domain question answering, and Contriever learns dense representations without relevance supervision \citep{bm25,karpukhin-etal-2020-dense,contriever}. Query-side enhancement offers another route to a better candidate pool. Query2doc expands a query with an LLM-generated pseudo-document, while HyDE retrieves through the embedding of a hypothetical document \citep{wang-etal-2023-query2doc,gao-etal-2023-precise}. These approaches improve candidate acquisition, but their immediate decision remains which evidence enters the pool and how it is initially ordered.

The candidate pool matters because generation is conditioned on the evidence ultimately supplied to the model. RAG jointly marginalizes over retrieved documents, Fusion-in-Decoder integrates information from multiple passages, and REPLUG demonstrates retrieval augmentation for frozen black-box language models \citep{rag,izacard-grave-2021-leveraging,shi-etal-2024-replug}. Together, these results show that evidence availability and quality can materially affect downstream answers. Yet the value of retrieval is query-dependent: non-parametric memory is especially useful for long-tail and knowledge-intensive questions, while some frequently observed facts can be answered from parametric memory \citep{mallen-etal-2023-trust}. Evidence selection must therefore consider not only whether a passage can be retrieved, but also whether it is useful to the generator for the current query.

\subsection{Relevance-Oriented Neural and LLM-Based Reranking}
Reranking refines an acquired candidate pool, most commonly by estimating query--document relevance. Supervised neural rerankers increase interaction capacity beyond first-stage retrieval: BERT reranking cross-encodes query--passage pairs, ColBERT uses contextualized late interaction, and sequence-to-sequence rankers such as monoT5 and RankT5 formulate ranking through generated relevance signals or ranking losses \citep{nogueira2020passagererankingbert,colbert,nogueira-etal-2020-document,t5}. These architectures differ in efficiency and interaction granularity, but they learn an ordering objective from relevance supervision or relevance-oriented training data.

Zero-shot LLM rerankers replace task-specific ranking training with prompted or likelihood-based relevance estimation. YesNo asks an LLM for a binary relevance judgment, QLM scores how likely a query is under a document, and RankGPT elicits a listwise ordering through prompting \citep{yesno,qlm,rankgpt}. RankVicuna extends listwise reranking to open-source instruction-tuned models \citep{pradeep2023rankvicuna}. Their scoring mechanisms vary, but the decision target remains principally the relevance of a candidate to the query.

Some work aligns ranking more directly with answer generation. RankRAG instruction-tunes one model to perform both context ranking and retrieval-augmented generation \citep{yu2024rankrag}. This joint objective connects evidence ordering with downstream use, but it requires additional model training. More broadly, relevance and generator-side usefulness are related without being equivalent. A relevant passage can repeat information the generator already represents or introduce distracting evidence, while a lower-ranked passage can resolve an ambiguity and increase semantic answer stability. This gap motivates a complementary signal that evaluates a document through its effect on generator behavior rather than by estimating another relevance score.

\subsection{Semantic Consistency and Black-Box Confidence Estimation}
LLM confidence estimation includes verbalized self-assessment and agreement across sampled outputs. Verbalized methods ask a model to state whether it knows an answer or to assess the likely correctness of a response. Language models can exhibit useful self-evaluation under suitable elicitation formats \citep{kadavath2022language}, but black-box confidence reports can remain miscalibrated or overconfident \citep{xiong2024can}. Their reliability consequently depends on the model's ability to express its own confidence, not only on the evidence contained in its generated answers.

Sampling-based methods instead examine variation across repeated generations. Self-consistency samples multiple reasoning paths and aggregates their answers \citep{wang2023selfconsistency}. Semantic entropy further groups outputs by meaning, separating surface-form variation from disagreement over the underlying answer \citep{farquhar2024detecting}. For black-box use, the resulting semantic concentration is informative because it can be estimated from generated text without logits or hidden states.

CAR draws on this semantic-consistency perspective but changes the unit of analysis. Rather than interpreting an absolute confidence value as document quality, it contrasts query-only and document-conditioned conditions for the same query. The contrast asks whether adding a candidate makes sampled answer meanings more concentrated or more dispersed relative to the query-specific control. Semantic consistency thereby becomes a generator-side document-usefulness signal.

\subsection{Generator-Guided Evidence Intervention in RAG}
Generator-side signals have also been used to decide how a RAG system should intervene on evidence. The intervention target differs across methods. Self-RAG trains reflection tokens that govern retrieval, generation and critique, while FLARE initiates forward-looking retrieval when forthcoming generation is judged insufficiently supported \citep{asai2024selfrag,jiang2023flare}. These methods primarily decide whether or when retrieval should occur. CRAG instead assesses retrieval quality and invokes corrective actions when the acquired evidence is unreliable \citep{yan2024crag}.

Evidence-aware RAG extends intervention beyond a single retrieval trigger. Recent work coordinates internal and external knowledge, prunes or selects evidence for generator use, and adjusts retrieval--generation policies according to model feedback \citep{ipm-4,ipm-5,song2026igp}. Such systems move beyond static retrieve-then-generate pipelines by allowing generator-side assessment to influence evidence acquisition, selection, fusion or answer production.

This literature establishes the value of closing the loop between the generator and external evidence, but most interventions operate at the level of system actions or broader retrieval--generation policies. A distinct problem remains after a candidate ranking already exists: how to use generator feedback to correct that order without discarding the ranking information supplied by the retriever or reranker. This local intervention scope is the point at which CAR complements adaptive and corrective RAG.

\subsection{Positioning of CAR}
CAR lies at the intersection of three lines of work. Relevance-oriented retrieval and reranking provide an initial candidate order; semantic consistency provides an observable generator-side stability signal; and generator-guided evidence intervention motivates using that signal to alter evidence exposure. Confidence-aware reranking such as LLM-Confidence Reranker is adjacent in its use of sampled-answer consistency \citep{song2026lcr}, while CAR specifically uses a query-conditioned change in semantic stability to constrain correction of an existing order.

\Cref{tab:method_positioning} compares these approaches along five dimensions: approach family, representative methods, primary signal, intervention and treatment of the prior order, and task-training requirements. The distinction is consequential: first-stage retrievers construct the candidate order, relevance rerankers infer a new relevance order, and adaptive or corrective RAG systems typically change retrieval or generation actions. CAR instead operates after these components have produced a top-$n$ ranking.

Within that scope, CAR is a training-free, black-box and evidence-constrained minimum-perturbation correction method. It treats query-conditioned semantic-stability change as a document-usefulness signal, induces coarse precedence constraints from that evidence and preserves the baseline order wherever those constraints do not require a reversal. CAR therefore complements rather than replaces relevance-based ranking. Its applicability depends on the candidate set containing meaningfully different evidence and on sampled answers yielding a discriminative semantic-stability signal; gains can diminish when the baseline order is already strong, the candidates are uniformly uninformative or semantic equivalence cannot be judged reliably.

\begin{table*}[t]
\centering
\caption{\textbf{Positioning of CAR among Retrieval and Reranking Approaches.} The approach families are compared by representative methods, primary signal, intervention and treatment of the prior order, and task-training requirements. Unlike methods that construct or replace a ranking or alter broader RAG actions, CAR uses query-conditioned stability change to correct an existing order while preserving within-bin precedence and requiring no task-specific training.}
\label{tab:method_positioning}
\input{wrappers/table2_method_positioning_en}
\end{table*}

\section{Methodology}
\label{sec:methodology}

\subsection{Problem Formulation and Design Objective}
Given a query $q$, a baseline retriever or reranker $\psi$ returns a top-$n$ candidate order
\[
\pi_q^{\psi}=[d_1,\ldots,d_n].
\]
We treat $\pi_q^{\psi}$ as a strict total order. If two candidates have equal baseline scores, their order in the ranker's actual output provides deterministic tie-breaking. CAR receives this ordered list and the candidate texts and returns a corrected order $\widehat{\pi}_q$. It neither requires the numerical baseline scores nor changes or retrains the baseline ranker or generator.

The baseline and CAR address different estimation targets. A relevance-oriented ranker orders documents by lexical matching, representation similarity, learned relevance or prompted relevance judgments. CAR instead asks how a document changes the generator's semantic answer stability under the same query. These targets can disagree: a topically relevant passage may introduce ambiguity or reinforce an unstable interpretation, whereas a lower-ranked passage may make the sampled answers converge on one meaning. CAR therefore estimates a document's marginal generator-side usefulness.

This distinction determines the design objective. Generator evidence should introduce only the pairwise precedence relations that it clearly supports. Among all rankings satisfying those relations, the method should retain as many baseline pairwise preferences as possible. CAR is thus an evidence-constrained minimum-perturbation rank correction method: confidence changes induce coarse precedence constraints, and CAR returns the feasible ranking with minimum Kendall distance from the baseline order. Kendall distance is appropriate because CAR requires only an ordinal candidate list and because each counted inversion corresponds directly to one baseline pairwise preference that the intervention overturns.

The formulation yields four design principles. First, usefulness must be measured relative to a query-specific control because baseline answer stability varies across queries. Second, small changes from finite sampling should be treated as indeterminate rather than actionable. Third, generator evidence should constrain only coarse movement classes instead of replacing the baseline with a noisy continuous score. Fourth, once those constraints are satisfied, every remaining baseline relation should be preserved. The following subsections derive the signal, constraints and closed-form rank correction from these principles.

\begin{center}
\begin{minipage}{\textwidth}
\centering
\captionof{table}{\textbf{Notation Used in CAR.} Definitions of the main symbols used in the CAR formulation and rank-correction procedure.}
\label{tab:notation}
\input{wrappers/table1_notation_en}
\end{minipage}
\end{center}

\Cref{fig:car_framework} gives the overall workflow, and the following subsections define each component formally.

\subsection{Semantic Stability of Sampled Answers}
For an input condition $x$, stochastic generation induces a latent distribution over answer meanings,
\[
P_{\phi}(z\mid x),
\]
where $z$ denotes semantic content rather than a surface string. The condition $x$ may contain the query alone or the same query paired with a candidate document. At the conceptual level, the concentration
\[
c_{\phi}(x)=\max_z P_{\phi}(z\mid x)
\]
describes how strongly the generator's answer distribution concentrates on its dominant meaning. High concentration means repeated answers tend to express one meaning; low concentration means that probability mass is distributed across competing meanings.

The latent distribution is an explanatory object, not an input available to CAR. Operationally, CAR samples $k$ answers $A_{\phi}(x)=\{a_1,\ldots,a_k\}$ and partitions them into empirical semantic clusters $\widehat{\mathcal Z}_x$. Its stability estimate is
\[
\widehat c_{\phi}(x)
=
\max_{C\in\widehat{\mathcal Z}_x}\frac{|C|}{k}.
\]
Clustering at the meaning level avoids treating lexical paraphrases as disagreement. The resulting largest-cluster proportion is observable from sampled outputs and can therefore serve as a black-box indicator of semantic stability without access to logits or hidden states.

It measures concentration among sampled answer meanings; it neither measures answer accuracy nor estimates the probability that an answer is correct. A generator can repeatedly produce the same incorrect meaning, while a genuinely ambiguous query can support several defensible meanings. CAR consequently uses $\widehat c_{\phi}(x)$ only to compare the stability induced by two input conditions, never as a standalone correctness score.

Because $\widehat c_{\phi}(x)$ is computed from $k$ samples, it is discrete in increments of $1/k$ and is sensitive to sampling and clustering variation near a decision boundary. This observation motivates an indifference margin before any rank movement is allowed. The concrete sampling and semantic-clustering protocol is specified in \cref{sec:experimental_setup}; the rank-correction formulation requires only the resulting empirical clusters.

\subsection{Confidence Change as a Controlled Contrast}
CAR uses the query-only input as a control condition and defines the document-level signal as
\[
\widehat\Delta_{\phi}(q,d)
=
\widehat c_{\phi}(q,d)-\widehat c_{\phi}(q).
\]
Both terms concern the same query; they differ only in whether the candidate document is supplied. The contrast therefore characterizes the document's marginal contribution to sampled-answer stability relative to the generator's query-specific starting point.

Absolute document-conditioned stability is insufficient for this purpose. An easy query may have high $\widehat c_{\phi}(q,d)$ for nearly every document simply because $\widehat c_{\phi}(q)$ is already high. Conversely, a difficult query may remain only moderately concentrated after receiving a document even when that document creates a substantial improvement over the query-only condition. Ranking by $\widehat c_{\phi}(q,d)$ alone would confound these query-specific baselines with document usefulness. The within-query contrast removes this particular confound and asks the narrower question relevant to CAR: does the document add stability beyond what the generator already exhibits for this query?

A positive $\widehat\Delta_{\phi}(q,d)$ denotes incremental stabilization, a negative value denotes destabilization or interference, and a value near zero provides insufficient directional evidence at the available sampling budget. This is a controlled contrast and a generator-side usefulness signal; no causal interpretation is assumed.

\subsection{Evidence-Constrained Intervention by QT and CM}
The controlled contrast should not alter every ranking. The query threshold (QT) is a query-level intervention gate:
\[
\widehat c_{\phi}(q)\ge T_q
\quad\Longrightarrow\quad
\widehat\pi_q=\pi_q^{\psi}.
\]
When query-only stability is already high, the room for useful correction is limited and acting on noisy document-level differences creates a greater risk of overcorrection. The gate therefore preserves the baseline exactly and avoids all document-conditioned evaluations for that query. Only when $\widehat c_{\phi}(q)<T_q$ does CAR collect document-conditioned evidence.

For an opened gate, write $\widehat\Delta_i=\widehat\Delta_{\phi}(q,d_i)$. When $m>0$, the confidence margin (CM) converts each controlled contrast into a three-level evidence label:
\[
b_i=
\begin{cases}
+1, & \widehat\Delta_i\ge m,\\
0,  & -m<\widehat\Delta_i<m,\\
-1, & \widehat\Delta_i\le -m.
\end{cases}
\]
The labels $+1$, $0$ and $-1$ mean promote, preserve and demote, respectively. For $m>0$, CM defines a non-empty indifference region around the query-only reference. Documents whose confidence changes fall inside this region are preserved, whereas changes reaching the positive or negative boundary trigger promotion or demotion.

When $m=0$, the indifference region vanishes and the evidence label is instead
\[
b_i=
\begin{cases}
+1, & \widehat\Delta_i\ge 0,\\
-1, & \widehat\Delta_i<0.
\end{cases}
\]
Following the implementation's ordered if--else rule, non-negative confidence changes are assigned to the promote bin and negative changes to the demote bin; consequently, the preserve bin is empty. Setting $m=0$ therefore removes the confidence-margin buffer. The assignment of a zero change to the promote bin is a deterministic boundary convention retained for exact consistency with the implementation, rather than an interpretation of zero change as positive evidence.

CM is a finite-sampling indifference margin, not a statistical confidence interval. It creates an operational buffer in which small changes are treated as insufficient evidence because they may result from finite sampling, semantic-clustering variation or nearly equivalent meanings. A direct continuous reranking by $\widehat\Delta_{\phi}$ would force distinctions even inside this buffer. In contrast, the three labels retain only the direction of changes large enough to cross the chosen margin.

QT and CM thus impose two levels of evidence constraint. QT determines whether any intervention is warranted for the query. Conditional on intervention, CM determines which cross-document precedence classes are supported. The next subsection shows that the original order-preserving binning rule is exactly the minimum-change projection of the baseline order onto these constraints.

\subsection{Minimum-Perturbation Rank Correction}
Let $r_{\sigma}(d)$ be the position of document $d$ in permutation $\sigma$, and let $S_n$ denote all permutations of the $n$ candidates. The evidence labels define the feasible set
\[
\mathcal F_q(b)
=
\left\{
\sigma\in S_n:
b_i>b_j
\Longrightarrow
r_{\sigma}(d_i)<r_{\sigma}(d_j)
\right\}.
\]
Feasibility requires every promoted document to precede every preserved or demoted document and every preserved document to precede every demoted document. It deliberately imposes no ordering constraint on documents with the same label.

Because candidates are indexed by their strict baseline order, the unnormalized Kendall distance from a permutation $\sigma$ to $\pi_q^{\psi}$ is
\[
d_K(\sigma,\pi_q^{\psi})
=
\sum_{1\le i<j\le n}
\mathbf 1\!\left[r_{\sigma}(d_i)>r_{\sigma}(d_j)\right].
\]
Each term records whether $\sigma$ reverses one pair preferred by the baseline. CAR can therefore be characterized as
\[
\widehat\pi_q
=
\arg\min_{\sigma\in\mathcal F_q(b)}
d_K(\sigma,\pi_q^{\psi}).
\]
The objective is lexicographic in interpretation: generator evidence first determines the admissible cross-bin precedence constraints; among rankings that satisfy them, CAR overturns the fewest baseline pairwise preferences. The minimum is asserted only within $\mathcal F_q(b)$, not over other ranking objectives or unconstrained notions of relevance.

Let $B^{+}$, $B^{0}$ and $B^{-}$ be the subsequences obtained by scanning $\pi_q^{\psi}$ and appending documents with labels $+1$, $0$ and $-1$, respectively.

\noindent\textbf{Proposition 1 (Minimum-perturbation characterization).}
If $\pi_q^{\psi}$ is a strict total order, the stable concatenation
\[
B^{+}\oplus B^{0}\oplus B^{-}
\]
is the unique permutation in $\mathcal F_q(b)$ with minimum Kendall distance from $\pi_q^{\psi}$.

\noindent\emph{Proof sketch.}
For every pair with different labels, feasibility fixes the higher-label document before the lower-label document; any resulting reversal of their baseline order is therefore unavoidable. Documents within one bin have no additional evidence constraint. Reversing any such pair adds a Kendall inversion without satisfying a new constraint, whereas retaining its baseline order adds none. The minimum consequently preserves every within-bin baseline relation. Since the baseline order is strict, each bin has one such order, and their forced cross-bin order yields the unique stable concatenation above. \hfill$\square$

This characterization makes five properties explicit. \emph{Identity}: if the QT gate does not open, the output equals the baseline. \emph{Evidence precedence}: a document with a higher label always precedes one with a lower label. \emph{Within-bin preservation}: equally labelled documents retain their baseline relative order. \emph{Idempotence}: for a fixed candidate set and fixed confidence decisions, applying CAR again does not change the result. \emph{Decision stability}: deterministic perturbations that cross neither the QT boundary nor any $\pm m$ boundary leave the output unchanged.

\subsection{Algorithm and Complexity Boundary}
\input{wrappers/algorithm_car_en}

Algorithm 1 separates stability estimation from rank correction. It scans candidates in baseline order and appends each one to exactly one ordered bin. Its promote-first, demote-second and preserve-last branches implement both definitions above: they yield three bins for $m>0$ and the stated two-bin boundary behavior for $m=0$. The returned concatenation is therefore the closed-form solution in Proposition 1; no generic permutation solver and no continuous sorting by $\widehat\Delta_{\phi}$ are used. Once the labels are known, constructing the corrected ranking takes $O(n)$ time and $O(n)$ auxiliary space.

The computational cost should distinguish generated samples, semantic judgments and rank construction. For candidate depth $n$ and sampling budget $k$, the number of generated answer samples is
\[
N_{\mathrm{sample}}
=
k\left[
1+n\,\mathbf 1\!\left\{\widehat c_{\phi}(q)<T_q\right\}
\right].
\]
The first term is always incurred for the query-only condition; the second is incurred only when the QT gate opens. These are generated samples, not necessarily separate API requests, because implementations may batch multiple samples or conditions.

Semantic clustering adds meaning judgments for each evaluated condition. Their number depends on the configured clustering procedure; CAR supports comparison reuse and batching. In practice, generator sampling and semantic judgments dominate the linear-time minimum-Kendall projection. These stages can be parallelized across candidates, but their total workload still grows with candidate depth and sampling budget.

This profile defines CAR's deployment boundary. It is intended for top-$n$ correction after a short candidate list has been retrieved, not for first-stage search over an entire corpus. It is most suitable when repeated generator sampling is affordable and the semantic clustering procedure distinguishes answer meanings reliably. The QT gate narrows the cost by bypassing document-conditioned evaluation when the query-only answer distribution is already sufficiently stable.

\section{Experiments}
\label{sec:experiments}

We first define the common experimental protocol in
\cref{sec:experimental_setup}, and then evaluate ranking effectiveness,
component contributions, operating robustness and downstream answer-quality
consistency through four research questions:
\begin{itemize}
\item \textbf{RQ1 (\cref{sec:main_results}):} Does CAR improve ranking quality across the evaluated
tasks and baseline ranking methods?
\item \textbf{RQ2 (\cref{sec:rq2}):} Which components enable CAR to improve ranking quality
while limiting perturbations to the baseline order?
\item \textbf{RQ3 (\cref{sec:rq3}):} How robust are CAR's gains across operating settings,
retriever and generator families, and sampling budgets?
\item \textbf{RQ4 (\cref{sec:rq4}):} Are CAR's ranking gains aligned with
downstream answer-quality gains?
\end{itemize}

\subsection{Experimental Setup}
\label{sec:experimental_setup}

\subsubsection{Datasets and Evaluation Tasks}
We use NQ for open-domain question answering, HotpotQA for multi-hop question
answering, and FEVER for fact checking
\citep{nq,yang-etal-2018-hotpotqa,fever}. These BEIR tasks vary in
query form, corpus size and relevance density. \Cref{tab:dataset_stats}
summarizes their evaluation populations. For every run, queries in each dataset
are randomly shuffled and split 1:9 into validation and test. This yields
validation/test counts of 345/3,107 for NQ, 740/6,665 for HotpotQA and
666/6,000 for FEVER. We repeat the random validation/test split procedure 10
times and report results over these 10 repeated runs.
\begin{table}[t]
\centering
\caption{\textbf{Statistics of the Evaluated BEIR Tasks.} The columns report the task, number of evaluation queries, corpus size and mean number of relevant documents per query.}
\label{tab:dataset_stats}
\input{wrappers/table3_dataset_stats_en}
\end{table}

For the downstream check in RQ4, we construct a fixed NQ-answerable (NQ-a)
subset by combining short answers from the original NQ release with relevance
judgments from BEIR-NQ. Because the two sources do not provide a ready-to-use
joint mapping, we manually align their queries and retain the 2,728 cases for
which both annotations are available. NQ-a is therefore a fixed aligned subset,
rather than the full NQ evaluation population, and is used only for downstream
evaluation rather than parameter selection.

\subsubsection{Retrieval and Reranking Baselines}
BM25 and Contriever represent sparse lexical and unsupervised dense retrieval
\citep{bm25,contriever}. Each provides a top-10 candidate list. We preserve the
retriever-only order and additionally evaluate YesNo, QLM, RankGPT, ColBERT,
Cross-Encoder and RankT5 \citep{yesno,qlm,rankgpt,colbert,
nogueira2020passagererankingbert,t5}. The first three are zero-shot LLM
rerankers; the last three are supervised neural rerankers.

\subsubsection{Generator and Semantic-Judge Models}
The main setting uses Qwen2.5-7B-Instruct (Qwen) for answer sampling and
semantic judgment \citep{qwen2.5}. Generator-family robustness additionally
uses GLM-4-9B-Chat (GLM) and InternLM2.5-7B-Chat (InternLM)
\citep{glm2024chatglm,cai2024internlm2}. Together with Qwen, these form the
three evaluated generator families. The same family supplies sampling and
semantic judgments within a condition, and CAR requires only generated text.

\subsubsection{Prompts and Semantic Clustering}
Answer sampling uses the following template:
\begin{grayquote}
\texttt{Context: \{context\}}\\
\texttt{Question: \{question\}}\\
\texttt{Answer:}
\end{grayquote}
The query-only control leaves context empty. Pairwise semantic judgments use
the fixed prompt:
\begin{grayquote}
\texttt{We are evaluating answers to the question "\{question\}"}\\
\texttt{Here are two possible answers:}\\
\texttt{Possible Answer 1: \{text1\}}\\
\texttt{Possible Answer 2: \{text2\}}\\
\texttt{Does Possible Answer 1 semantically entail Possible Answer 2? Respond
with only one of: entailment, contradiction, or neutral.}
\end{grayquote}
Two answers share a cluster when neither directional judgment is contradiction
and the pair is not neutral in both directions. Confidence is the largest
semantic-cluster proportion, not correctness probability.

\subsubsection{Sampling and Decoding Settings}
The main experiment uses $k=10$ answers per input, temperature 1.0 and a
100-token maximum. Semantic judgments use temperature 0 and a 20-token
maximum. The sampling-budget analysis retains $k=1$ as the no-CAR baseline
reference and evaluates CAR at $k=2,\ldots,10$.

\subsubsection{Downstream Generation Protocol}
For RQ4, we use BM25 retrieval and the same seven baseline ranking methods as
in the main evaluation. In each repeated run, CAR reuses the operating
parameters selected on the corresponding NQ validation split; no parameter is
tuned on NQ-a. Baseline and CAR start from the same candidate pool and differ
only in document order, and the downstream generator receives the top five
documents under each resulting order. We hold the query, candidate documents,
prompt, generator and decoding configuration fixed between the paired
conditions. Answer generation uses Qwen under the same configuration as the
main experiment, so the comparison isolates the downstream effect of CAR's
rank correction.

\subsubsection{Evaluation Metrics}
NDCG@$k$ measures early-ranking quality with linear gain:
\[
\mathrm{DCG}@k=\sum_{i=1}^{k}\frac{\mathrm{rel}_i}{\log_2(i+1)},
\qquad
\mathrm{NDCG}@k=\frac{\mathrm{DCG}@k}{\mathrm{IDCG}@k}.
\]
We use NDCG@5 as the primary ranking metric because the top-ranked evidence is
most consequential to a downstream generator. For readability, all NDCG@5
scores and absolute differences are reported on a 0--100 scale. Relative
improvement is
\[
\Delta\%=\frac{\mathrm{Score}_{\mathrm{CAR}}-
\mathrm{Score}_{\mathrm{Baseline}}}{\mathrm{Score}_{\mathrm{Baseline}}}
\times100.
\]
Normalized Kendall distance (nKD) measures realized perturbation from the
baseline order:
\[
\mathrm{nKD}(\pi,\pi_0)=
\frac{d_K(\pi,\pi_0)}{\binom{|\pi_0|}{2}},
\]
which normalizes pairwise-order reversals for comparison across rankings. NQ-a
generation transfer uses token-level
$F_1=2PR/(P+R)$, where $P$ and $R$ are token-overlap precision and recall
against the reference answer; this directly measures downstream answer quality.

\subsection{Main Results (RQ1)}
\label{sec:main_results}
Using BM25 as the common first-stage retriever, RQ1 evaluates CAR across the
three tasks and seven baseline ranking methods. \Cref{tab:bm25_main} reports
these results; transfer to the dense Contriever retriever is evaluated
separately in RQ3.

\begin{center}
\begin{minipage}{\textwidth}
\centering
\captionof{table}{\textbf{Main Results with BM25.} Scores are mean NDCG@5 $\pm$ sample standard deviation over 10 repeated runs, and $\Delta\%$ is the average percentage change from the corresponding baseline; bold and underlined means mark the best and second-best results in each column, and AVG averages the three datasets. CAR is non-negative in all 21 displayed dataset--ranker cells and gives the largest average relative gain to YesNo (+29.9\%), whereas the supervised neural rankers improve by at most 0.1\% on average.}
\label{tab:bm25_main}
\input{wrappers/table4_bm25_main_results_en}
\end{minipage}
\end{center}

\textbf{CAR improves or preserves ranking effectiveness, with no observed
degradation across the evaluated settings.} Averaging the CAR-minus-baseline difference equally across the
21 cells gives a gain of +0.70 NDCG@5 points; averaging the corresponding
cell-level relative changes gives +5.53\%. Based on the unrounded means, 12
cells improve and the remaining nine are unchanged; no evaluated cell shows a
negative mean change.

\textbf{NQ shows the broadest and most consistent improvements.} CAR improves
all seven NQ ranker settings. On HotpotQA and FEVER, one of seven and four of
seven settings improve, respectively. All HotpotQA and FEVER cells for the
three supervised neural rerankers remain unchanged at the reported precision.
This pattern is consistent with CAR making selective corrections when
generator-side feedback complements the baseline ranking.

\textbf{CAR provides substantially larger gains
for some zero-shot LLM rerankers than for the supervised neural rerankers.}
The largest improvement occurs for YesNo, with an average relative gain of
+29.9\% and an absolute gain of +3.94 NDCG@5 points. However, the zero-shot
results are heterogeneous: the average relative gains are +29.9\% for YesNo,
+0.7\% for QLM and +4.0\% for RankGPT, compared with at most +0.1\% for each
supervised neural reranker. The result therefore primarily reflects the large
improvement for YesNo rather than a uniform advantage across zero-shot
methods.

\subsection{Component and Rank-Correction Analysis (RQ2)}
\label{sec:rq2}
RQ2 examines the distinct roles of CAR's three correction mechanisms: QT gates
intervention at the query level, CM converts document-level confidence changes
into evidence classes, and order-preserving binning projects those classes onto
the baseline order. Because an effectiveness gain obtained through many
pairwise reversals differs operationally from the same gain obtained with a
smaller correction, we report relative NDCG@5 gain together with nKD.
\Cref{fig:rq2_components,fig:ranker_fullres} report the component-level and
ranker-level comparisons, respectively. In \cref{fig:rq2_components}, datasets
form the columns and retrievers form the rows; each panel pairs effectiveness
with realized perturbation under the corresponding validation-selected
operating point.
\begin{center}
\begin{minipage}{\textwidth}
\centering
\input{wrappers/component_ablation_effect_perturbation_en}
\captionof{figure}{\textbf{Component Ablation: Effectiveness and Perturbation.} Columns are NQ, HotpotQA and FEVER, and rows are BM25 and Contriever. In each panel, the upper axis reports paired NDCG@5 change from the baseline for w/o QT, w/o CM and Full CAR over 10 repeated runs: bars and error bars show the mean and SD, points show ranker--run deltas with negative values in red, and annotations give the negative share. The lower axis reports nKD under the corresponding validation-selected operating point, with shorter lollipops indicating fewer reversals. Removing QT often increases both gain and perturbation and produces less consistent ranker--run deltas on HotpotQA/BM25. The CM=0 variant has the lowest realized nKD but a smaller gain. Full CAR has non-negative panel-level mean gains in all six settings.}
\label{fig:rq2_components}
\end{minipage}
\end{center}
\begin{center}
\begin{minipage}{\textwidth}
\centering
\input{wrappers/ranker_fullres_car_comparison_en}
\captionof{figure}{\textbf{CAR versus Full-Resolution Sorting.} Panel (a) reports paired CAR-minus-full-resolution NDCG@5 differences by ranker; diamonds and horizontal intervals show the mean over 10 repeated runs and descriptive 95\% Student-t interval after equal averaging over three datasets and two retrievers, small marks show the six dataset--retriever means, and W/T/L is from CAR's perspective. Panel (b) reports $\mathrm{nKD}(\mathrm{Full\mbox{-}resolution})-\mathrm{nKD}(\mathrm{CAR})$. Full-resolution sorting reuses the paired CAR QT, uses no CM and undergoes no separate parameter selection. Its aggregate nKD is about 2.1 times CAR's, while effectiveness differences are smaller and depend on the ranker.}
\label{fig:ranker_fullres}
\end{minipage}
\end{center}

\textbf{QT acts as a query-level safeguard against unnecessary intervention.}
In the w/o QT variant, the gate is always open and CM is validation-selected.
This variant produces the largest gain and perturbation on several
dataset--retriever panels. On NQ, w/o QT and Full CAR both reach a +3.6\%
relative gain, but their nKD values are 0.094 and 0.070, respectively. The
additional intervention is not uniformly reliable: on HotpotQA/BM25, 85.7\%
of the ranker--run w/o QT deltas are non-positive even though the panel mean
remains positive. By contrast, Full CAR has a non-negative panel-level mean in
all six dataset--retriever settings. These results support QT's intended role:
it can avoid additional reversals when opening the gate provides no aggregate
benefit and improves directional consistency in the least favorable panel.

\textbf{CM controls the evidence resolution after the gate opens.} The w/o CM
variant fixes CM=0 and selects QT normally. Under the promote-first boundary,
zero change is assigned to promote and the preserve bin is empty, reducing the
three evidence classes to a sign-based two-bin partition. Empirically, this
variant has the lowest nKD in all six panels but also a smaller attainable
gain. The lower nKD arises because the two-bin partition preserves more
within-bin baseline pairs even as CM=0 removes the indifference region. Full CAR
uses the three-bin rule to distinguish strong changes from near-zero changes, obtaining
additional effectiveness at the cost of some additional reversals. Thus,
evidence selectivity and realized nKD have a non-monotonic relationship.

\textbf{Order-preserving correction sharply reduces perturbation at a modest
aggregate effectiveness cost.} Averaged over all settings, full-resolution
sorting yields a +3.44\% relative gain at nKD 0.090, whereas CAR yields +2.63\%
at 0.043. The 0.81-point difference in relative gain therefore accompanies
about 2.1 times as much pairwise-order perturbation under full-resolution
sorting. The effectiveness difference is also
ranker-dependent: the full-resolution mean advantage is most pronounced for
YesNo, whereas the other rankers are closer to parity or favor CAR in some
dataset--retriever settings. Panel (b) nevertheless shows greater perturbation
from full-resolution sorting for every ranker.

\textbf{Effectiveness and perturbation reveal a clear operating trade-off.}
Mean differences describe effect magnitude, while W/T/L summarizes directional
consistency across the six dataset--retriever settings after run averaging.
Together, the two RQ2 figures clarify the role of each component: QT determines
whether to intervene, CM sets the resolution of the document-level evidence
classes, and stable bin concatenation preserves within-class order relative to
full-resolution confidence sorting. NDCG@5 and nKD capture distinct objectives.
We therefore report both. This
empirical trade-off complements the minimum-Kendall characterization in
\cref{sec:methodology}, which concerns optimality under CAR's bin constraints.

\subsection{Robustness across Operating and System Settings (RQ3)}
\label{sec:rq3}
Whereas RQ2 isolates the contributions of CAR's components, RQ3 examines the
robustness of its gains across QT/CM operating profiles, sparse and dense
retriever families, generator families and sampling budgets.
The QT/CM profiles in \cref{fig:qtcm_profiles}, cell-level retriever results in
\cref{tab:bm25_main} and \cref{tab:contriever_main} of \appref{app:contriever},
retriever aggregates in \cref{tab:retriever_robustness}, and generator-family
and sampling-budget comparisons in
\cref{fig:generator_family,fig:sampling_budget} provide the corresponding
evidence.

\subsubsection{QT/CM Operating Robustness}
\begin{center}
\begin{minipage}{\textwidth}
\centering
\input{wrappers/qt_cm_profile_sensitivity_en}
\captionof{figure}{\textbf{Robustness to QT and CM.} Mean test-set NDCG@5 gains of CAR over Baseline remain non-negative across all profiled QT and CM values on NQ, HotpotQA and FEVER, with broad positive regions indicating limited sensitivity to either parameter. Shaded bands denote descriptive 95\% Student-t intervals over 10 repeated runs.}
\label{fig:qtcm_profiles}
\end{minipage}
\end{center}
\textbf{Broad QT/CM regions support robustness.} Across the one-dimensional
profiles, NQ, HotpotQA and FEVER retain non-negative gains over QT 0--1 and CM
0--1, with broad positive regions rather than a single isolated point.

\subsubsection{Retriever Robustness}
\textbf{CAR improves ranking effectiveness under both BM25 and Contriever.}
Mean relative gains are +5.53\% under BM25 and
+4.64\% under Contriever. Positive paired means occur in
10 of 21
dataset--ranker cells for Contriever and
12 of 21 for BM25; all
14 of 14 NQ cells are positive. The
complete Contriever and BM25 tables provide the cell-level evidence for these
aggregates, and the retriever summary reports the same pattern. CAR is therefore
not tied to either a sparse or dense retrieval family.
\begin{center}
\begin{minipage}{0.56\textwidth}
\centering
\captionof{table}{\textbf{Robustness across Retrievers.} Baseline and +CAR NDCG@5 are averaged over three datasets, seven rankers and 10 repeated runs for each retriever. CAR improves both sparse BM25 (+5.53\%) and dense Contriever (+4.64\%), showing that the gains are not tied to one retrieval family.}
\label{tab:retriever_robustness}
\input{wrappers/table9_retriever_en}
\end{minipage}
\end{center}

\textbf{Variation across rankers is larger than variation between retriever
families.} YesNo receives average relative gains of
+29.9\% with BM25 and +28.6\% with
Contriever, QLM changes by +0.7\% and
+3.0\%, respectively, and every retriever-level average
gain for ColBERT, Cross-Encoder and RankT5 is at most
+0.2\%; many HotpotQA and FEVER cells remain
unchanged at displayed precision. Because this ranker-dependent spread recurs
under both retrievers, the variation is associated more with the baseline
method than with the retriever family.

\subsubsection{Generator-Family Robustness}
\begin{center}
\begin{minipage}{\textwidth}
\centering
\input{wrappers/generator_family_robustness_en}
\captionof{figure}{\textbf{Robustness across Generator Families.} Each cell reports paired mean CAR-minus-Baseline NDCG@5 for one generator and dataset, averaged over 10 repeated runs, two retrievers and seven rankers. All nine generator--dataset aggregates are positive, and InternLM produces the largest gain on each dataset, with an overall mean gain of +1.78 NDCG@5 points.}
\label{fig:generator_family}
\end{minipage}
\end{center}
\textbf{CAR is effective across generator families, with generator-dependent
gain magnitudes.} All nine displayed
generator--dataset aggregates are positive. Across datasets, the mean gains are
+0.96 for Qwen, +1.11 for GLM and +1.78 NDCG@5 points for InternLM, with
InternLM producing the largest gain on NQ, HotpotQA and FEVER. Each cell in the
generator--dataset matrix averages over 10 repeated runs, two retrievers and
seven rankers. The consistently positive direction supports robustness across
the evaluated generator families, whereas the variation in gain magnitude
shows that generator choice still matters. In practice, the generator used by
CAR should therefore be validated for the target task, because a better-matched
model may provide a more discriminative stability signal and yield larger
ranking improvements.

\subsubsection{Sampling-Budget Sensitivity}
\begin{center}
\begin{minipage}{\textwidth}
\centering
\input{wrappers/sampling_budget_sensitivity_en}
\captionof{figure}{\textbf{Sensitivity to the Sampling Budget.}
On NQ, $k=1$ is the no-CAR reference, while $k=2,\ldots,10$ use
separately validation-selected CAR operating points. BM25 and Contriever
are shown in separate facets with one curve per ranker. CAR yields a
positive mean gain at every evaluated budget (+0.48 to +0.87 NDCG@5
points), with an overall upward trend as the sampling budget increases.}
\label{fig:sampling_budget}
\end{minipage}
\end{center}
\textbf{CAR maintains positive gains across all evaluated sampling budgets,
with an overall upward trend as $k$ increases.} Taking $k=1$ as the no-CAR
reference, the separately validation-selected CAR operating points for
$k=2,\ldots,10$ yield mean paired gains of +0.48 to +0.87 NDCG@5 points.
Despite local fluctuations, ranking gains generally increase with the sampling
budget.

\subsection{Consistency between Ranking and Downstream Answer-Quality Gains (RQ4)}
\label{sec:rq4}
Ranking gains alone do not establish that a corrected evidence order benefits
answer generation. RQ4 therefore evaluates the baseline and CAR orders on the
fixed NQ-a subset under the controlled downstream protocol described in
\cref{sec:experimental_setup}. For each of the seven ranking methods, we compare
token-level F1 against the reference answers over the same 10 repeated runs.
\Cref{fig:generation_consistency} reports the corresponding ranking and
answer-quality gains.
\begin{center}
\begin{minipage}{\textwidth}
\centering
\input{wrappers/ranking_to_generation_en}
\captionof{figure}{\textbf{Consistency between Ranking and Answer-Quality Gains.} On the fixed 2,728-query NQ-a subset, each point represents one ranking method; the horizontal and vertical axes report mean relative NDCG@5 and token-level F1 gains over 10 repeated runs, with descriptive 95\% Student-t intervals (df=9). The method-level relative ranking and answer-quality gains are strongly aligned ($\rho=0.93$).}
\label{fig:generation_consistency}
\end{minipage}
\end{center}

\textbf{On NQ-a, CAR's ranking gains are accompanied by higher answer quality
and are strongly aligned with answer-quality gains across rankers.} Mean
token-level F1 increases from 14.74 for the baseline order to 15.17 under CAR,
a paired gain of +0.43 points, and the 95\% interval is above zero for five of
the seven rankers. Across the seven method-level means, relative NDCG@5 and
token-level F1 gains have Spearman $\rho=0.93$, indicating that rankers with
larger CAR ranking gains also tend to show larger answer-quality gains. These
results support downstream consistency between CAR's ranking and generation
benefits on NQ-a and agree with prior evidence that the quality and ordering of
supplied evidence can affect RAG generation
\citep{rag,izacard-grave-2021-leveraging,
shi-etal-2024-replug,yu2024rankrag}.

\subsection{Case Study}
\label{sec:case_study}
\Cref{fig:case_study} traces a query from NQ. The relevant Himadri document is promoted from rank 8 to 1 after a positive confidence change. A near-neutral document is preserved and moves only from rank 1 to 2 as other bins are assembled, whereas the negatively scored Antarctica document falls from rank 4 to 10. The resulting top-five order replaces unsupported evidence with the ground-truth relevant passage, raising query-level NDCG@5 from 0 to 1. The example illustrates the intended coarse action: promote clear supporting evidence, retain near-neutral evidence, and demote destabilizing evidence without reordering documents inside a bin.
\begin{center}
\begin{minipage}{\textwidth}
\centering
\input{wrappers/within_query_car_correction_en}
\captionof{figure}{\textbf{Case Study.} CAR reranking for a query from NQ. The relevant Himadri document is promoted from rank 8 to 1 after a positive confidence change; a near-neutral document moves only from rank 1 to 2 as the bins are assembled; and the negatively scored Antarctica document falls from rank 4 to 10. The resulting top-five ranking replaces unsupported evidence with the ground-truth relevant passage, raising query-level NDCG@5 from 0 to 1 while preserving document order within each bin.}
\label{fig:case_study}
\end{minipage}
\end{center}

\subsection{Takeaways}
\label{sec:takeaways}
\textbf{RQ1:} CAR consistently improves or preserves early-ranking quality in
the BM25-centered evaluation, with particularly broad gains on NQ and the
largest benefits for some zero-shot relevance judgments.
\textbf{RQ2:} The component analyses show that CAR's effectiveness and
conservatism arise jointly from query-level gating, confidence-margin-based
evidence classes and order-preserving binning. Together, these mechanisms
avoid unnecessary intervention and limit the extra perturbation introduced by
full-resolution sorting.
\textbf{RQ3:} CAR remains effective across operating profiles,
sparse and dense retrievers, generator families and sampling budgets. The
positive pattern spans thresholds and system configurations.
\textbf{RQ4:} On NQ-a, improvements in ranking are reflected in downstream
answer quality, and rankers that benefit more in ranking also tend to benefit
more in generation. The case study illustrates the same mechanism at the
query level: CAR promotes clearly supporting evidence, preserves near-neutral
evidence and demotes destabilizing passages while retaining within-class order.

\section{Conclusion}
\label{sec:conclusion}
\subsection{Summary of Contributions and Findings}
This work addresses the relevance--usefulness gap in RAG evidence ordering.
Query--document relevance remains essential for acquiring and ranking
candidates, but it does not fully determine whether a passage will help the
downstream generator answer the current query. Evidence ordering can therefore
combine relevance with a complementary assessment of how each candidate
affects generator behavior.

CAR operationalizes this assessment through a controlled contrast between
query-only and document-conditioned semantic answer stability. The resulting
query-conditioned signal estimates generator-side usefulness. CAR uses this
feedback to correct rather than reconstruct an existing ranking: coarse
evidence constraints authorize selected precedence changes, and the
feasible ranking with minimum Kendall distance preserves every baseline
preference that those constraints do not require it to overturn.

Across the evaluated retrieval, ranking and generation settings, the results
support the feasibility and robustness of this generator-aware, conservative
rank-correction paradigm. They further indicate that improved evidence ordering
and downstream generation benefit are consistent in the controlled setting
examined here.

\subsection{Implications}
\subsubsection{Theoretical Implications}
The formulation extends query--document ranking with a second relation: how the
query and a candidate jointly affect generator behavior. Evidence usefulness in
RAG is therefore conditioned on the downstream generator and is not fully
determined by relevance. The objectives remain distinct but complementary:
relevance supplies an informative prior order, while generator-side usefulness
can identify cases in which the effect of evidence warrants correction.

The results also motivate a less strictly one-way view of the RAG pipeline.
Beyond retrieval, reranking and generation as a linear sequence, observable
generator behavior can provide feedback for evidence organization. CAR closes
this loop locally by applying a controlled contrast after candidate ranking to
guide limited correction. Within the evaluated scope,
generator feedback can thus serve as a control signal for evidence organization.

Finally, the minimum-perturbation formulation gives rank correction a broader
interpretation. An evidence-constrained intervention changes only justified
pairwise relations and preserves the remaining preferences. CAR's
minimum-Kendall correction realizes this principle: preserve informative priors
and intervene only where complementary evidence supports change.

\subsubsection{Practical Implications}
At the system level, CAR can be deployed as a post-ranking, generator-aware
correction layer after a sparse or dense retriever or an existing reranker. It
requires neither retraining of the baseline ranker or generator nor access to
internal model states. The available correction space naturally depends on how
much useful ordering information the baseline already contains: generator
feedback can support larger changes when deficiencies remain, while a stronger
baseline leaves fewer relations that warrant reversal.

Deployment should therefore balance three objectives rather than optimize a
ranking metric in isolation: downstream effectiveness, the cost of generator
sampling and semantic judgments, and the perturbation introduced into the
baseline order. CAR makes this trade-off explicit through selective
intervention and minimum perturbation. More generally, RAG evidence optimization
can be viewed as a constrained intervention problem in which effectiveness gains
must be weighed against computational cost and disruption of an informative
prior ranking.

\subsection{Limitations}
First, CAR inherits the limitations of semantic stability as a behavioral
signal. Concentration among sampled answer meanings is neither a calibrated
measure of correctness or factuality nor a relevance probability. A generator
can be stably wrong, while a genuinely ambiguous query can produce several
legitimate meanings. The signal also remains sensitive to sampling variation
and the reliability of semantic-equivalence judgments.

Second, CAR has a computational boundary. The cost of repeated generation and
semantic judgments grows with candidate depth and sampling budget, limiting
the method to short top-$n$ correction rather than corpus-scale first-stage
retrieval.

Finally, CAR is bounded by both its candidate pool and its document-wise
modeling. It cannot recover relevant evidence that is absent from the candidate
pool, nor does its controlled contrast systematically capture complementarity,
redundancy, conflict or other interactions within an evidence set.

\subsection{Future Work}
These limitations motivate three directions for future work. First, improving
signal validity requires generator-side usefulness measures that combine
stability with factual verification, faithfulness or evidential support,
together with uncertainty-aware grouping and more reliable
semantic-equivalence judgments. These advances could better distinguish stable
errors from supported answers.

Second, reducing inference cost calls for adaptive or boundary-aware sampling,
early stopping, caching, batching and efficient semantic grouping, so that
computation is concentrated near decision boundaries.

Finally, moving beyond document-wise correction requires modeling
complementarity, redundancy, conflict and other multi-document interactions in
larger, noisier or incomplete candidate pools. This line of work can further
investigate whether generator feedback supports evidence selection, pruning or
acquisition beyond CAR's current local reranking scope.

\appendix
\section{Contriever Results for Retriever Robustness}
\label{app:contriever}
\begin{center}
\begin{minipage}{\textwidth}
\centering
\captionof{table}{\textbf{Retriever-Robustness Results with Contriever.} Scores are mean NDCG@5 $\pm$ sample standard deviation over 10 repeated runs, and $\Delta\%$ is the average percentage change from the corresponding baseline; bold and underlined means mark the best and second-best results in each column, and AVG averages the three datasets. CAR is non-negative in all 21 displayed dataset--ranker cells and gives the largest average relative gain to YesNo (+28.6\%), whereas the supervised neural rankers improve by at most 0.2\% on average.}
\label{tab:contriever_main}
\input{wrappers/tableA1_contriever_main_results_en}
\end{minipage}
\end{center}

%% file: wrappers/fig1_framework_overview_en.tex
\includegraphics[width=0.93\textwidth]{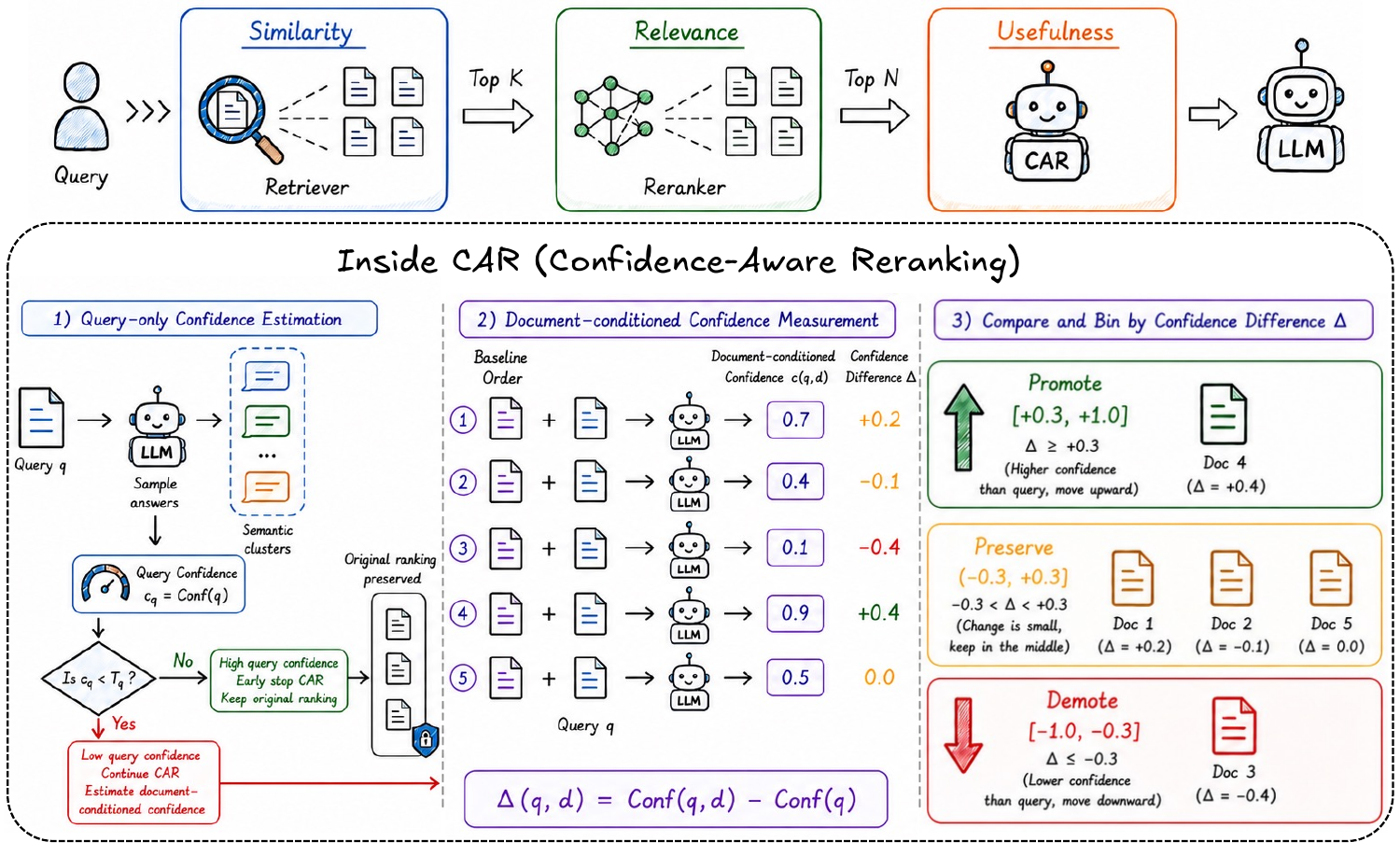}

%% file: wrappers/table2_method_positioning_en.tex
\scriptsize
\setlength{\tabcolsep}{3pt}
\begin{tabular}{p{0.15\textwidth}p{0.16\textwidth}p{0.19\textwidth}p{0.31\textwidth}p{0.09\textwidth}}
\toprule
Approach & Examples & Primary signal & Intervention and prior order & Task training \\
\midrule
First-stage retrieval & BM25, Contriever & Lexical / representation matching & Constructs the initial candidate order & No \\
Supervised reranking & ColBERT, Cross-Encoder, RankT5 & Learned relevance & Replaces the prior order & Yes \\
Zero-shot LLM reranking & YesNo, QLM, RankGPT & Prompted relevance / query likelihood & Replaces the prior order & No \\
Adaptive/corrective RAG & Self-RAG, FLARE, CRAG & Self-evaluation / retrieval quality & Changes retrieval or generation actions & Varies \\
CAR & This work & Query-conditioned stability change & Minimum-perturbation correction; preserves within-bin order & No \\
\bottomrule
\end{tabular}

%% file: wrappers/table1_notation_en.tex
\begingroup
\footnotesize
\setlength{\tabcolsep}{3pt}
\renewcommand{\arraystretch}{0.98}
\begin{tabular}{p{0.22\textwidth}p{0.70\textwidth}}
\toprule
Symbol & Meaning \\
\midrule
$\psi$ & Baseline ranker (retriever or reranker) \\
$\phi$ & Generator (large language model) \\
$q$ & Query \\
$d_i$ & Candidate document at baseline rank $i$ \\
$n$ & Number of candidate documents \\
$k$ & Number of samples per input \\
$\pi_q^{\psi}$ & Strict candidate order returned by $\psi$ \\
$\widehat{\pi}_q$ & Candidate order after CAR correction \\
$A_{\phi}(x)$ & Set of $k$ sampled answers under input condition $x$ \\
$P_{\phi}(z\mid x)$ & Latent distribution over semantic answer meanings $z$ \\
$\widehat{\mathcal Z}_x$ & Empirical semantic partition of $A_{\phi}(x)$ \\
$c_{\phi}(x)$ & Conceptual semantic concentration $\max_z P_{\phi}(z\mid x)$ \\
$\widehat c_{\phi}(x)$ & Empirical largest-cluster proportion under $x$ \\
$\widehat\Delta_{\phi}(q,d_i)$ & Controlled contrast $\widehat c_{\phi}(q,d_i)-\widehat c_{\phi}(q)$ \\
$T_q$ & Query threshold (QT) \\
$m$ & Confidence margin (CM) \\
$b_i$ & Evidence label for $d_i$ in $\{+1,0,-1\}$ \\
$B^{+},B^{0},B^{-}$ & Ordered promote, preserve and demote bins \\
$S_n$ & Set of all permutations of the $n$ candidates \\
$r_{\sigma}(d)$ & Position of document $d$ in permutation $\sigma$ \\
$\mathcal F_q(b)$ & Permutations satisfying all evidence-induced precedence constraints \\
$d_K(\sigma,\pi_q^{\psi})$ & Unnormalized Kendall distance from the baseline order \\
\bottomrule
\end{tabular}
\endgroup

%% file: wrappers/algorithm_car_en.tex
\begingroup
\footnotesize
\setlength{\tabcolsep}{4pt}
\renewcommand{\arraystretch}{1.02}
\begin{tabular}{p{0.94\textwidth}}
\toprule
\textbf{Procedure} $\textsc{Estimate-Stability}(x,\phi,k)$ \\
\midrule
E1. Sample $k$ answers $A_{\phi}(x)=\{a_1,\ldots,a_k\}$ from generator $\phi$ under input condition $x$. \\
E2. Apply the semantic-clustering procedure to $A_{\phi}(x)$ and obtain the partition $\widehat{\mathcal Z}_x$. \\
E3. Return $\widehat c_{\phi}(x)=\max_{C\in\widehat{\mathcal Z}_x}|C|/k$. \\
\midrule
\textbf{Procedure} $\textsc{Car-Rerank}(q,\pi_q^{\psi},\phi,k,T_q,m)$ \\
\textbf{Input:} query $q$; strict baseline order $\pi_q^{\psi}=[d_1,\ldots,d_n]$; generator $\phi$; sample count $k$; QT $T_q$; CM $m\ge0$. \\
\textbf{Output:} corrected order $\widehat\pi_q$. \\
\midrule
C1. Set $\widehat c_q\leftarrow\textsc{Estimate-Stability}(q,\phi,k)$. \\
C2. If $\widehat c_q\ge T_q$, return $\pi_q^{\psi}$. \\
C3. Initialize ordered lists $B^{+}\leftarrow[\ ]$, $B^{0}\leftarrow[\ ]$ and $B^{-}\leftarrow[\ ]$. \\
C4. For $i=1,\ldots,n$ in baseline order: \\
\quad a. Set $\widehat c_i\leftarrow\textsc{Estimate-Stability}((q,d_i),\phi,k)$ and $\widehat\Delta_i\leftarrow\widehat c_i-\widehat c_q$. \\
\quad b. If $\widehat\Delta_i\ge m$, set $b_i\leftarrow+1$ and append $d_i$ to $B^{+}$. \\
\quad c. Else if $\widehat\Delta_i\le-m$, set $b_i\leftarrow-1$ and append $d_i$ to $B^{-}$. \\
\quad d. Else set $b_i\leftarrow0$ and append $d_i$ to $B^{0}$. \\
C5. Return $\widehat\pi_q\leftarrow B^{+}\oplus B^{0}\oplus B^{-}$. \\
\bottomrule
\end{tabular}
\endgroup

%% file: wrappers/table3_dataset_stats_en.tex
\small
\resizebox{\columnwidth}{!}{%
\begin{tabular}{llrrr}
\toprule
Dataset & Task & \#Query & \#Corpus & Avg. Rel. D/Q \\
\midrule
NQ & Question answering & 3,452 & 2,681,468 & 1.20 \\
HotpotQA & Multi-hop question answering & 7,405 & 5,233,329 & 2.00 \\
FEVER & Fact checking & 6,666 & 5,416,568 & 1.20 \\
\bottomrule
\end{tabular}}

%% file: wrappers/table4_bm25_main_results_en.tex
\scriptsize
\setlength{\tabcolsep}{2.6pt}
\renewcommand{\arraystretch}{1.05}
\resizebox{\textwidth}{!}{%
\begin{tabular}{lcccccccc}
\toprule
\multirow{2}{*}{\textbf{Method}} & \multicolumn{2}{c}{NQ} & \multicolumn{2}{c}{HotpotQA} & \multicolumn{2}{c}{FEVER} & \multicolumn{2}{c}{AVG} \\
 & Score & $\Delta\%$ & Score & $\Delta\%$ & Score & $\Delta\%$ & Score & $\Delta\%$ \\
\midrule
\multicolumn{9}{l}{\textit{Retriever Only}} \\
\midrule
\textbf{BM25} & 4.92 {\tiny $\pm$ 0.08} & - & 54.37 {\tiny $\pm$ 0.12} & - & 31.21 {\tiny $\pm$ 0.13} & - & 30.17 & - \\
\quad +CAR & 5.37 {\tiny $\pm$ 0.10} & \textcolor{gaintext}{+9.2} & 54.37 {\tiny $\pm$ 0.12} & +0.0 & 32.06 {\tiny $\pm$ 0.13} & \textcolor{gaintext}{+2.7} & 30.60 & \textcolor{gaintext}{+4.0} \\
\midrule
\multicolumn{9}{l}{\textit{LLM-based Rerankers}} \\
\midrule
\textbf{YesNo} & 3.23 {\tiny $\pm$ 0.06} & - & 24.11 {\tiny $\pm$ 0.09} & - & 13.10 {\tiny $\pm$ 0.05} & - & 13.48 & - \\
\quad +CAR & 4.18 {\tiny $\pm$ 0.12} & \textcolor{gaintext}{+29.4} & 30.62 {\tiny $\pm$ 0.10} & \textcolor{gaintext}{+27.0} & 17.47 {\tiny $\pm$ 0.06} & \textcolor{gaintext}{+33.4} & 17.42 & \textcolor{gaintext}{+29.9} \\
\midrule
\textbf{QLM} & 6.13 {\tiny $\pm$ 0.11} & - & 50.38 {\tiny $\pm$ 0.07} & - & 36.04 {\tiny $\pm$ 0.12} & - & 30.85 & - \\
\quad +CAR & 6.25 {\tiny $\pm$ 0.15} & \textcolor{gaintext}{+1.9} & 50.38 {\tiny $\pm$ 0.07} & +0.0 & 36.10 {\tiny $\pm$ 0.13} & \textcolor{gaintext}{+0.2} & 30.91 & \textcolor{gaintext}{+0.7} \\
\midrule
\textbf{RankGPT} & 4.93 {\tiny $\pm$ 0.08} & - & 54.39 {\tiny $\pm$ 0.12} & - & 31.21 {\tiny $\pm$ 0.13} & - & 30.18 & - \\
\quad +CAR & 5.39 {\tiny $\pm$ 0.10} & \textcolor{gaintext}{+9.3} & 54.39 {\tiny $\pm$ 0.12} & +0.0 & 32.06 {\tiny $\pm$ 0.13} & \textcolor{gaintext}{+2.7} & 30.61 & \textcolor{gaintext}{+4.0} \\
\midrule
\multicolumn{9}{l}{\textit{Supervised Neural Rerankers}} \\
\midrule
\textbf{ColBERT} & 7.94 {\tiny $\pm$ 0.11} & - & 59.97 {\tiny $\pm$ 0.10} & - & 47.57 {\tiny $\pm$ 0.16} & - & 38.49 & - \\
\quad +CAR & 7.96 {\tiny $\pm$ 0.12} & \textcolor{gaintext}{+0.2} & 59.97 {\tiny $\pm$ 0.10} & +0.0 & 47.57 {\tiny $\pm$ 0.16} & +0.0 & 38.50 & \textcolor{gaintext}{+0.1} \\
\midrule
\textbf{Cross-Encoder} & 8.14 {\tiny $\pm$ 0.10} & - & \underline{61.94} {\tiny $\pm$ 0.08} & - & \underline{47.94} {\tiny $\pm$ 0.19} & - & 39.34 & - \\
\quad +CAR & 8.15 {\tiny $\pm$ 0.11} & \textcolor{gaintext}{+0.1} & \underline{61.94} {\tiny $\pm$ 0.08} & +0.0 & \underline{47.94} {\tiny $\pm$ 0.19} & +0.0 & 39.34 & \textcolor{gaintext}{+0.0} \\
\midrule
\textbf{RankT5} & \underline{8.33} {\tiny $\pm$ 0.11} & - & \textbf{62.95} {\tiny $\pm$ 0.08} & - & \textbf{48.77} {\tiny $\pm$ 0.18} & - & \underline{40.01} & - \\
\quad +CAR & \textbf{8.34} {\tiny $\pm$ 0.11} & \textcolor{gaintext}{+0.0} & \textbf{62.95} {\tiny $\pm$ 0.08} & +0.0 & \textbf{48.77} {\tiny $\pm$ 0.18} & +0.0 & \textbf{40.02} & \textcolor{gaintext}{+0.0} \\
\bottomrule
\end{tabular}}

%% file: wrappers/component_ablation_effect_perturbation_en.tex
\includegraphics[width=0.98\textwidth]{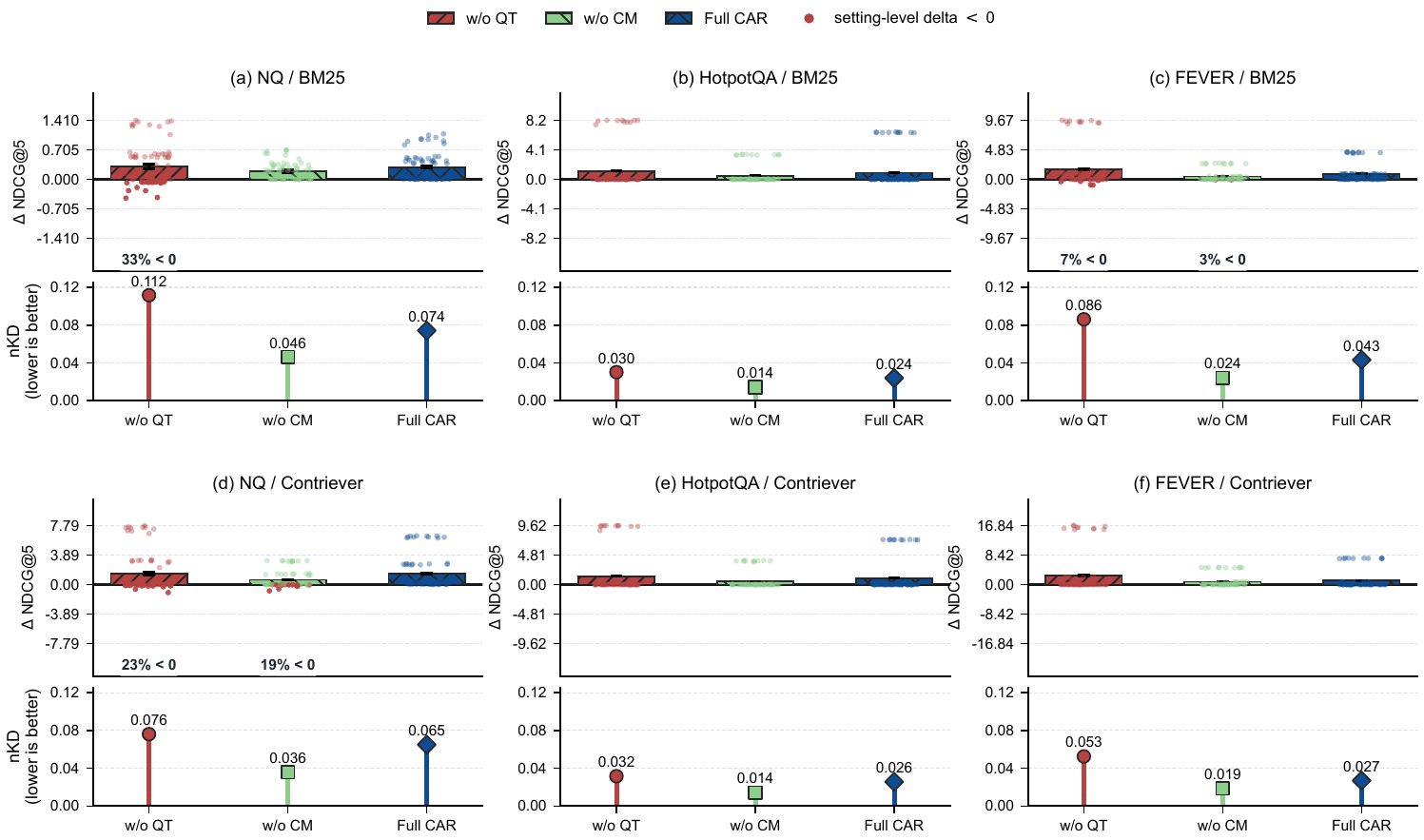}

%% file: wrappers/ranker_fullres_car_comparison_en.tex
\includegraphics[width=0.92\textwidth]{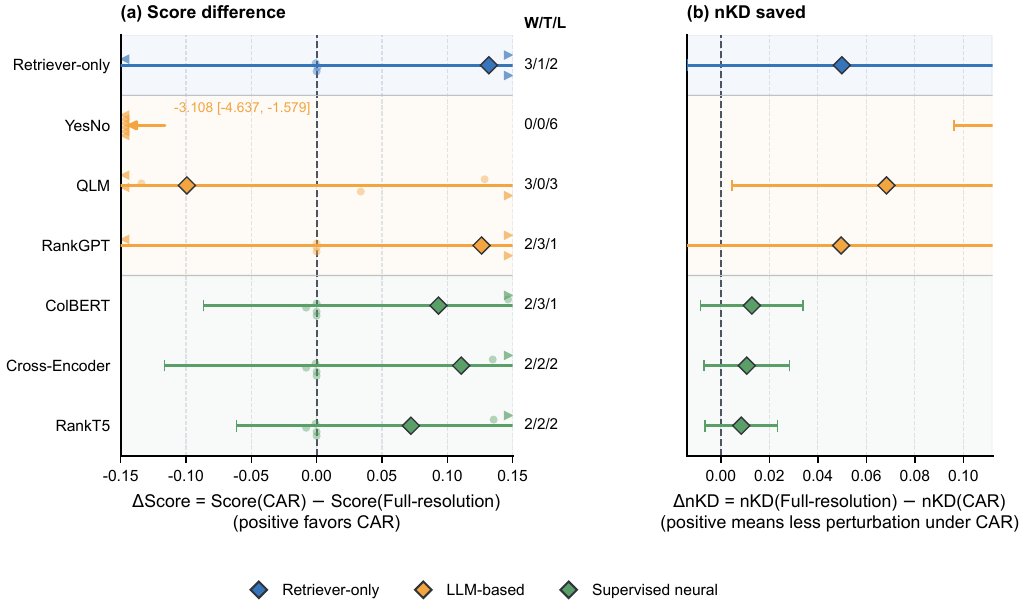}

%% file: wrappers/qt_cm_profile_sensitivity_en.tex
\includegraphics[width=0.98\textwidth]{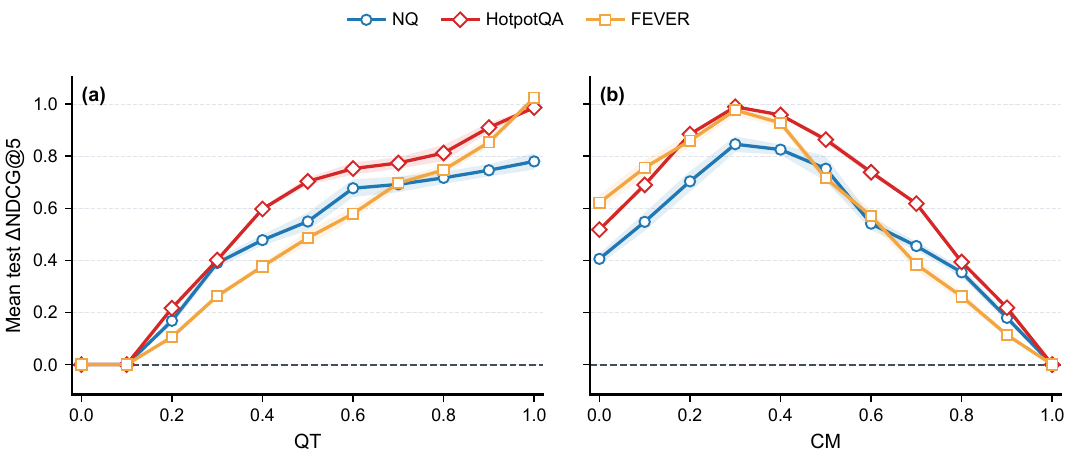}

%% file: wrappers/table9_retriever_en.tex
\small
\begin{tabular}{lccc}
\toprule
Retriever & Mean baseline & Mean +CAR & Mean $\Delta\%$ \\
\midrule
BM25 & 31.79 & 32.49 & \textcolor{gaintext}{+5.53} \\
Contriever & 55.17 & 56.39 & \textcolor{gaintext}{+4.64} \\
\bottomrule
\end{tabular}

%% file: wrappers/generator_family_robustness_en.tex
\includegraphics[width=0.98\textwidth]{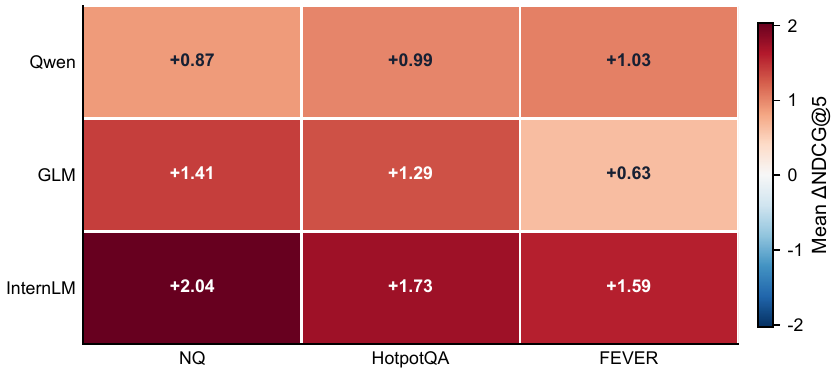}

%% file: wrappers/sampling_budget_sensitivity_en.tex
\includegraphics[width=0.92\textwidth]{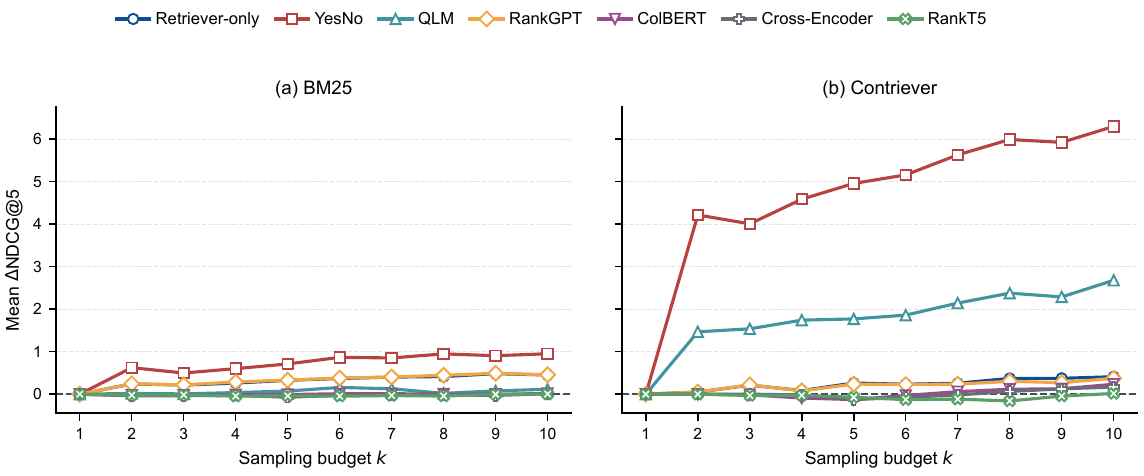}

%% file: wrappers/ranking_to_generation_en.tex
\includegraphics[width=0.92\textwidth]{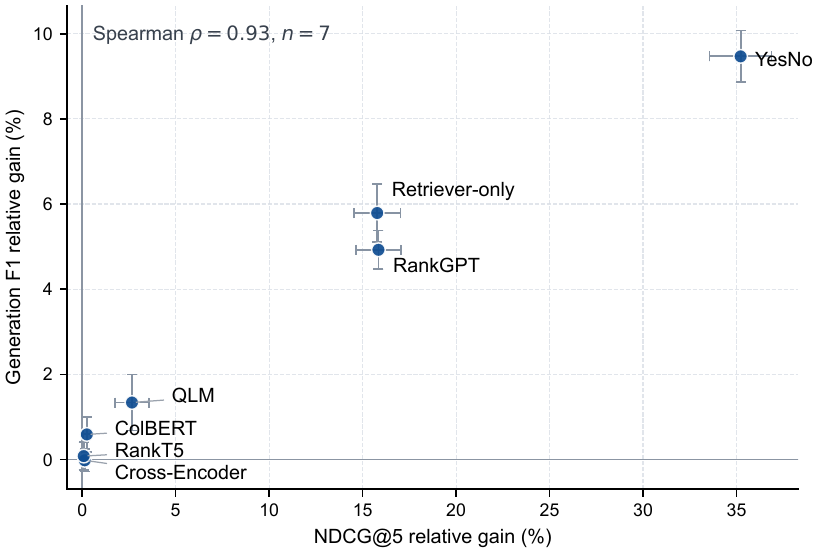}

%% file: wrappers/within_query_car_correction_en.tex
\includegraphics[width=0.92\textwidth]{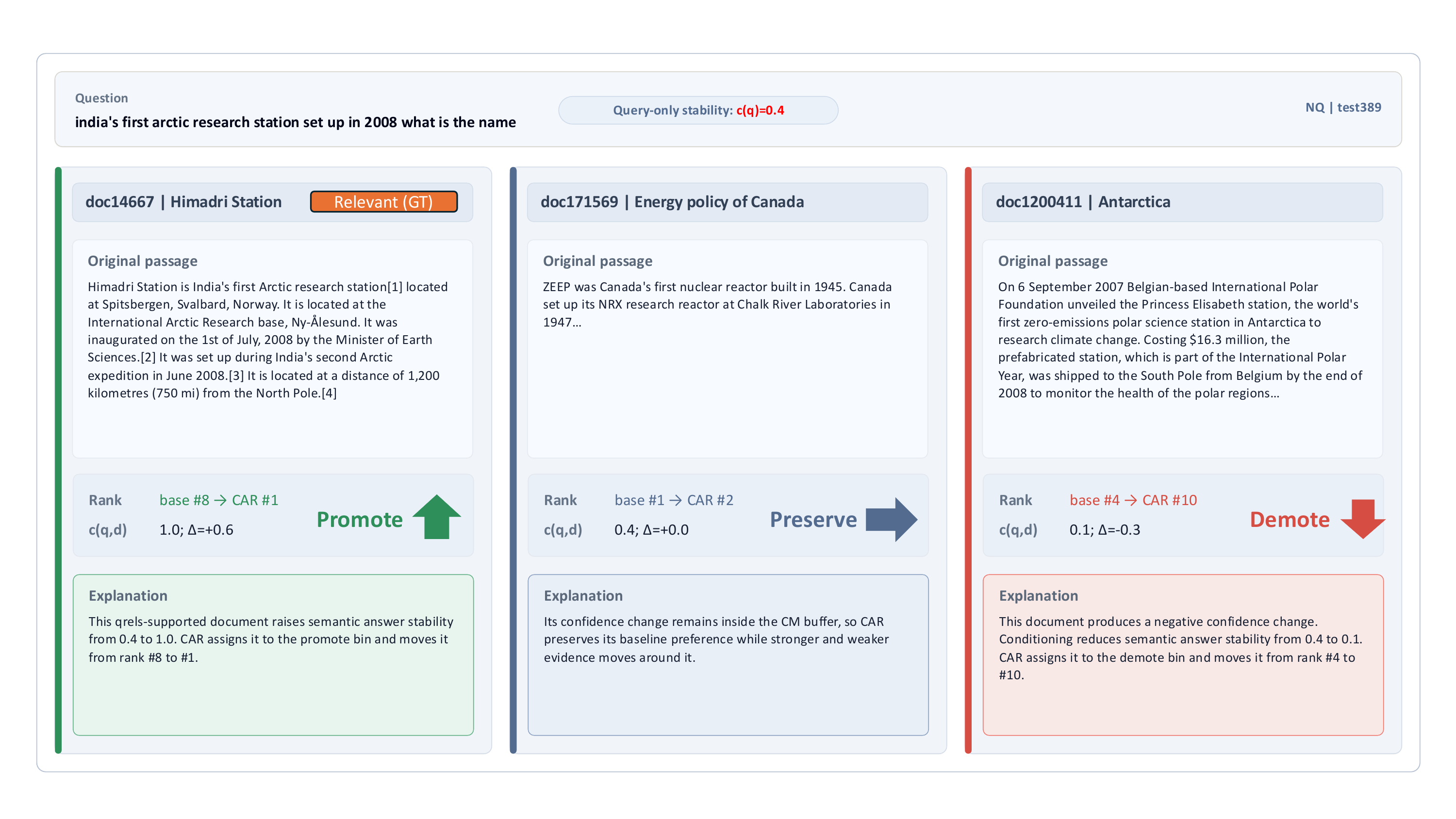}

%% file: wrappers/tableA1_contriever_main_results_en.tex
\scriptsize
\setlength{\tabcolsep}{2.6pt}
\renewcommand{\arraystretch}{0.90}
\resizebox{\textwidth}{!}{%
\begin{tabular}{lcccccccc}
\toprule
\multirow{2}{*}{\textbf{Method}} & \multicolumn{2}{c}{NQ} & \multicolumn{2}{c}{HotpotQA} & \multicolumn{2}{c}{FEVER} & \multicolumn{2}{c}{AVG} \\
 & Score & $\Delta\%$ & Score & $\Delta\%$ & Score & $\Delta\%$ & Score & $\Delta\%$ \\
\midrule
\multicolumn{9}{l}{\textit{Retriever Only}} \\
\midrule
\textbf{Contriever} & 45.46 {\tiny $\pm$ 0.21} & - & 60.45 {\tiny $\pm$ 0.09} & - & 74.35 {\tiny $\pm$ 0.11} & - & 60.09 & - \\
\quad +CAR & 45.87 {\tiny $\pm$ 0.21} & \textcolor{gaintext}{+0.9} & 60.45 {\tiny $\pm$ 0.09} & +0.0 & 74.35 {\tiny $\pm$ 0.11} & +0.0 & 60.23 & \textcolor{gaintext}{+0.3} \\
\midrule
\multicolumn{9}{l}{\textit{LLM-based Rerankers}} \\
\midrule
\textbf{YesNo} & 23.21 {\tiny $\pm$ 0.16} & - & 25.05 {\tiny $\pm$ 0.09} & - & 25.40 {\tiny $\pm$ 0.11} & - & 24.55 & - \\
\quad +CAR & 29.51 {\tiny $\pm$ 0.20} & \textcolor{gaintext}{+27.1} & 32.36 {\tiny $\pm$ 0.08} & \textcolor{gaintext}{+29.2} & 32.89 {\tiny $\pm$ 0.10} & \textcolor{gaintext}{+29.5} & 31.59 & \textcolor{gaintext}{+28.6} \\
\midrule
\textbf{QLM} & 34.88 {\tiny $\pm$ 0.19} & - & 54.80 {\tiny $\pm$ 0.13} & - & 55.73 {\tiny $\pm$ 0.11} & - & 48.47 & - \\
\quad +CAR & 37.56 {\tiny $\pm$ 0.21} & \textcolor{gaintext}{+7.7} & 54.80 {\tiny $\pm$ 0.13} & +0.0 & 56.47 {\tiny $\pm$ 0.17} & \textcolor{gaintext}{+1.3} & 49.61 & \textcolor{gaintext}{+3.0} \\
\midrule
\textbf{RankGPT} & 45.55 {\tiny $\pm$ 0.22} & - & 60.44 {\tiny $\pm$ 0.09} & - & 74.37 {\tiny $\pm$ 0.11} & - & 60.12 & - \\
\quad +CAR & 45.92 {\tiny $\pm$ 0.20} & \textcolor{gaintext}{+0.8} & 60.44 {\tiny $\pm$ 0.09} & +0.0 & 74.37 {\tiny $\pm$ 0.11} & +0.0 & 60.25 & \textcolor{gaintext}{+0.3} \\
\midrule
\multicolumn{9}{l}{\textit{Supervised Neural Rerankers}} \\
\midrule
\textbf{ColBERT} & 47.76 {\tiny $\pm$ 0.15} & - & 62.81 {\tiny $\pm$ 0.08} & - & 76.27 {\tiny $\pm$ 0.11} & - & 62.28 & - \\
\quad +CAR & 48.00 {\tiny $\pm$ 0.20} & \textcolor{gaintext}{+0.5} & 62.81 {\tiny $\pm$ 0.08} & +0.0 & 76.27 {\tiny $\pm$ 0.11} & +0.0 & 62.36 & \textcolor{gaintext}{+0.2} \\
\midrule
\textbf{Cross-Encoder} & 48.66 {\tiny $\pm$ 0.18} & - & \underline{65.45} {\tiny $\pm$ 0.09} & - & \underline{78.54} {\tiny $\pm$ 0.12} & - & 64.22 & - \\
\quad +CAR & 48.83 {\tiny $\pm$ 0.25} & \textcolor{gaintext}{+0.3} & \underline{65.45} {\tiny $\pm$ 0.09} & +0.0 & \underline{78.54} {\tiny $\pm$ 0.12} & +0.0 & 64.27 & \textcolor{gaintext}{+0.1} \\
\midrule
\textbf{RankT5} & \underline{50.98} {\tiny $\pm$ 0.18} & - & \textbf{66.75} {\tiny $\pm$ 0.08} & - & \textbf{81.63} {\tiny $\pm$ 0.11} & - & \underline{66.45} & - \\
\quad +CAR & \textbf{50.99} {\tiny $\pm$ 0.18} & \textcolor{gaintext}{+0.0} & \textbf{66.75} {\tiny $\pm$ 0.08} & +0.0 & \textbf{81.63} {\tiny $\pm$ 0.11} & +0.0 & \textbf{66.46} & \textcolor{gaintext}{+0.0} \\
\bottomrule
\end{tabular}}